# Real-time Video Target Tracking Algorithm Utilizing Convolutional Neural Networks (CNN)

Chaoyi Tan
Northeastern University
Boston, USA
tan.chaoyi@icloud.com

Xiangtian Li
University of California San Diego
San Diego, USA
xil160@ucsd.edu

Xiaobo Wang
Georgia Institute of Technology
Atlanta, USA
xwang468@gatech.edu

Zhen Qi
Northeastern University
Boston, USA
garyzhen79@gmail.com

Ao Xiang*
Northern Arizona University
Arizona, USA
*ax36@nau.edu

*Abstract*—This paper aims to research and implement a real-time video target tracking algorithm based on Convolutional Neural Networks (CNN), enhancing the accuracy and robustness of target tracking in complex scenarios. Addressing the limitations of traditional tracking algorithms in handling issues such as target occlusion, morphological changes, and background interference, our approach integrates target detection and tracking strategies. It continuously updates the target model through an online learning mechanism to adapt to changes in the target's appearance. Experimental results demonstrate that, when dealing with situations involving rapid motion, partial occlusion, and complex backgrounds, the proposed algorithm exhibits higher tracking success rates and lower failure rates compared to several mainstream tracking algorithms. This study successfully applies CNN to real-time video target tracking, improving the accuracy and stability of the tracking algorithm while maintaining high processing speeds, thus meeting the demands of real-time applications. This algorithm is expected to provide new solutions for target tracking tasks in video surveillance and intelligent transportation domains.

## I. INTRODUCTION

In the past few decades, the rapid advancement of computer science, notably AI, has underscored the significance of visual object tracking in computer vision [1]. This technology identifies and tracks targets across video frames, revealing its potential in various applications, from traffic management and surveillance to autonomous vehicles and medical imaging, where target tracking is pivotal [2]. However, practical applications encounter challenges, including scene complexity, background clutter, lighting changes, occlusion, and target shape variations, which may impair detection accuracy [3]. Small objects, due to limited pixels and information, are especially prone to oversight or misidentification in complex backgrounds. Additionally, unpredictable target motions demand robust and real-time tracking algorithms [4].

Traditional methods, relying on manually designed features like SIFT and SURF, excel in simple scenes but lack robustness and generalization in complex environments. The advent of deep learning, particularly CNN's breakthrough in image recognition, offers new solutions [5]. CNN, with its strong feature extraction and end-to-end learning, avoids manual feature design, learning effective features automatically [6]. This big data and complex model-based approach enhances object detection's accuracy and efficiency, becoming the mainstream algorithm.

CNN extracts high-level features through multi-layer convolution and pooling, exhibiting robustness to deformation and lighting changes, thus improving tracking accuracy. Learning directly from extensive data, CNN saves labor and adapts to diverse scenarios. End-to-end optimization further enhances system efficiency and performance[7]. With improved hardware and algorithms, CNN-based tracking now achieves real-time or near-real-time processing while maintaining high accuracy. Recent methods like Transformer-based tracking models struggle with real-time processing due to high computational costs. In contrast, our approach balances high accuracy with efficient computation.

This work draws upon insights from prior research, particularly in the application of deep reinforcement learning to dynamic environments, as demonstrated in [1]. The methodology outlined in their study offers valuable guidance, especially in terms of enhancing decision-making accuracy and adaptability in complex scenarios, which is highly relevant to improving the robustness of our CNN-based real-time tracking algorithm. The dual-channel attention mechanism proposed in [3] offers valuable insights for enhancing feature extraction, particularly when combining multiple data sources, such as RGB and NIR images. This approach significantly contributes to improving the segmentation accuracy and object integrity, which is particularly relevant for refining feature representation in our CNN-based real-time video tracking algorithm. This work draws on [6], particularly their CNN design and dropout use, which inform our approach to enhancing robustness and efficiency in real-time tracking.Insights from [8], particularly the integration of CNNs for spatial features and LSTMs for temporal dependencies, inspire enhancements to our real-time tracking framework's adaptability and robustness.

## II. METHODOLOGY

### A. Application of CNN in Real time Video Object Tracking

With the swift advancement of AI, particularly the proliferation of DL, video object tracking has seen unprecedented progress [8-9]. CNN, known for its robust feature extraction and efficiency, stands out in real-time tracking [10]. It automatically learns high-level features from raw images, mimicking the human visual system, which effectively represents shape, texture, and color—crucial for recognition and tracking. These CNN-extracted features outperform traditional ones, handling lighting changes, partial occlusion, and pose variations more adeptly [11]. Real-time tracking faces challenges such as rapid appearance changes, complex backgrounds, and target movement or occlusion, demanding quick response, accurate prediction, and robustness [12].

Introducing CNN offers new solutions, but inter-frame independent detection alone may miss continuous motion info, leading to tracking failure. Hence, this paper integrates correlated features with CNN for joint detection and tracking. Optical flow, describing pixel/feature point motion between frames, reflects target trajectories. Incorporating it into CNN provides temporal context, aiding in understanding dynamic changes. The network architecture (Figure 1) includes a Flow branch for processing optical flow, extracting feature maps that encapsulate motion and spatial correspondence between frames. The online learning mechanism in our framework continuously updates the target model in real-time, adapting to changes in target appearance and motion.

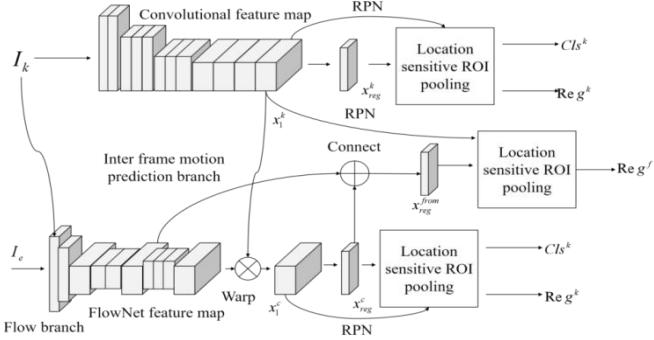

Fig. 1. Joint detection structure diagram combining correlation features and CNN

Through further analysis, the Flow branch predicts the target's next frame position, offering crucial prior knowledge for tracking algorithms. Its output trains an offset prediction module optimized by minimizing the offset loss [13-14]. During tracking, upon detecting the target, this module refines the position using the Flow branch's optical flow feature map, enhancing accuracy and stability. For real-time tracking, our framework is optimized in several ways: depthwise separable convolutions and bottleneck layers reduce parameters and complexity, while speeding up forward propagation [15]. Cross-layer connections, feature pyramids, and other strategies enable feature sharing and reuse, cutting down redundant calculations [16]. GPU acceleration and parallel computing asynchronously process video frames, boosting overall speed.

### B. Algorithm Design

CNN is a neural network constructed using a feedforward mechanism, whose core lies in gradually deepening the network structure through layer by layer convolution operations [17-18]. In the architecture of CNN, neurons serve as the fundamental processing units responsible for executing computational tasks. For a neuron with input dimension $x_1, x_2, x_3$, its output result is based on the operation between a specific convolution kernel (or filter) and the input data, which can be expressed in the following mathematical formula:

$$Y_{w,b}(x) = f(W^T x) = f\left(\sum_{i=1}^{3} W_i x_i + b_i\right) \quad (1)$$

Among them, $W_i$ represents the weight value of each node, and $b_i$ represents the bias constant. A neuron can be connected to other neurons to form a neural network.

In order to accurately predict the offset between frames to achieve the predetermined goal, this paper adopts a multi task loss function under the R-FCN framework [19]. This function is carefully designed with two core components: classification loss $L_{cls}$, responsible for precise classification; And single frame regression loss $L_{reg}$, used to optimize the target localization offset within a single frame. On this basis, we have made innovative extensions by introducing the inter frame regression loss $L_{frm}$, a new component specifically designed for training inter frame motion prediction branches, thereby further improving the performance of the model in cross frame object tracking [20]. For each training batch of $N$ candidate regions (RoIs), the model performs three parallel branches of prediction tasks: first, predicts the classification probability set $\{p_i\}_{i=1}^{N}$ for each RoI through the classification branch; Secondly, use the single frame regression branch to calculate the position offset set $\{b_i\}_{i=1}^{N}$ of each RoI within a single frame; Finally, through the inter frame regression branch, the model predicts the motion offset set $\{\Delta_i^{t+\tau}\}_{i=1}^{N}$ of each RoI between consecutive frames. The overall loss function is:

$$L(\{p_i\}, \{b_i\}, \{\Delta_i\}) = \frac{1}{N} \sum_{i=1}^{N} L_{cls}(p_i, p_i^*) +$$
$$\lambda \frac{1}{N_{fg}} \sum_{i=1}^{N} [p_i^* \geq 1] L_{reg}(b_i, b_i^*) + \quad (2)$$
$$\lambda \frac{1}{N_{cor}} \sum_{i=1}^{N} L_{frm}(\Delta_i^{t+\tau}, \Delta_i^{*,t+\tau})$$

In the process of learning coordinate parameterization for $b^*, \Delta^{*,t+\tau}$, we followed the classic coordinate parameterization strategy in R-CNN. Specifically, for a single target on the current frame $I_c$, its GroundTruth coordinate is represented as $B^{t+\tau} = (B_x^{t+\tau}, B_y^{t+\tau}, B_w^{t+\tau}, B_h^{t+\tau})$, where $B_x^{t+\tau}, B_y^{t+\tau}$ represents the horizontal and vertical coordinates of the center point of the target bounding box, and $B_w^{t+\tau}, B_h^{t+\tau}$ represents the width and height of the bounding box. The coordinates obtained by the model through prediction are represented as $P^{t+\tau} = (P_x^{t+\tau}, P_y^{t+\tau}, P_w^{t+\tau}, P_h^{t+\tau})$, and these predicted values also follow the format of bounding box center coordinates and dimensions, aiming to approximate the true GroundTruth coordinates. In addition, for the RoI coordinates generated by the RPN network on keyframe $I_k$, we denote them as $a^{t+\tau} = (a_x^{t+\tau}, a_y^{t+\tau}, a_w^{t+\tau}, a_h^{t+\tau})$. These RoI coordinates also contain the center point position and size information of the bounding box, and they play an important role in subsequent inter frame prediction and regression tasks.

$$\Delta_x^{t+\tau} = \frac{P_x^{t+\tau} - a_x^{t,t+\tau}}{a_x^{t,t+\tau}}, \Delta_x^{t+\tau} = \frac{P_y^{t+\tau} - a_y^{t+\tau}}{a_y^{t+\tau}} \quad (3)$$

$$\Delta_w^{t+\tau} = \log\left(\frac{P_w^{t+\tau}}{a_w^{t,t+\tau}}\right), \Delta_h^{t+\tau} = \log\left(\frac{P_h^{t+\tau}}{a_h^{t+\tau}}\right) \quad (4)$$

$$\Delta_x^{*,t+\tau} = \frac{B_x^{t+\tau} - a_x^{t,t+\tau}}{a_x^{t,t+\tau}}, \Delta_x^{t+\tau} = \frac{B_y^{t+\tau} - a_y^{t+\tau}}{a_y^{t+\tau}} \quad (5)$$

$$\Delta_w^{*,t+\tau} = \log\left(\frac{B_w^{t+\tau}}{a_w^{t,t+\tau}}\right), \Delta_h^{t+\tau} = \log\left(\frac{B_h^{t+\tau}}{a_h^{t+\tau}}\right) \quad (6)$$

In the inter frame motion prediction branch, we define two core steps for a specific foreground RoI: network forward prediction and final expected prediction. Specifically, equations (3) and (4) accurately describe the prediction results obtained by the RoI during the network transmission process. These two formulas capture the preliminary estimation of the current RoI's position or state changes in the next frame by the model, providing a foundation for subsequent fine-tuning and optimization. Subsequently, equations (5) and (6) represent the predicted results we expect to achieve for this RoI.

YOLOv3, a CNN based efficient object detection algorithm innovatively sets three anchor boxes for each grid cell, and further predicts three bounding boxes for each box. Each box accurately predicts four key values: $t_x, t_y$ represents position offset, and $t_w, t_h$ represents width and height [21]. If the center of the target falls in a certain cell and there is an offset $(c_x, c_y)$ relative to the upper left corner of the image, then these predicted values can be accurately adjusted using a specific formula to ensure the accuracy of target detection.

$$\begin{aligned} b_x &= \sigma(t_x) + c_x \\ b_y &= \sigma(t_y) + c_y \\ b_w &= p_w e^{t_w} \\ b_h &= p_h e^{t_h} \end{aligned} \quad (7)$$

Among them, $p_w, p_h$ represents the width and height of the anchor box corresponding to the grid.

III. RESULT ANALYSIS AND DISCUSSION

To verify the performance of the real-time video object tracking algorithm based on CNN in this paper, a comparative experiment is conducted between the algorithm proposed in this paper and the traditional SIFT based object tracking algorithm. Figure 2 intuitively and clearly presents the significant advantage of our algorithm in terms of target tracking success rate compared to traditional SIFT based target tracking algorithms.Our algorithm maintains an average FPS of 25 in dense environments with multiple moving targets, whereas traditional methods such as SIFT achieve only 15 FPS.While YOLOv5 excels in speed, our method demonstrates superior robustness in complex tracking scenarios, particularly under occlusion and rapid motion. This comparison not only highlights the powerful capabilities of modern DL technology in complex visual tasks, but also reveals the limitations of traditional feature descriptors in dealing with dynamically changing scenes [22-23]. Specifically, the reason why the algorithm in this article can achieve a higher tracking success rate is mainly attributed to its innovative combination of correlation features and the power of DL [24]. By automatically learning and extracting deep and abstract feature representations from images through CNN, algorithms can more effectively capture changes in the appearance, scale, and even partial occlusion of targets from different perspectives, thereby maintaining stable tracking of the targets. In addition, the introduction of correlated features further enhances the robustness of the algorithm in complex environments, making the tracking process more continuous and accurate [25].

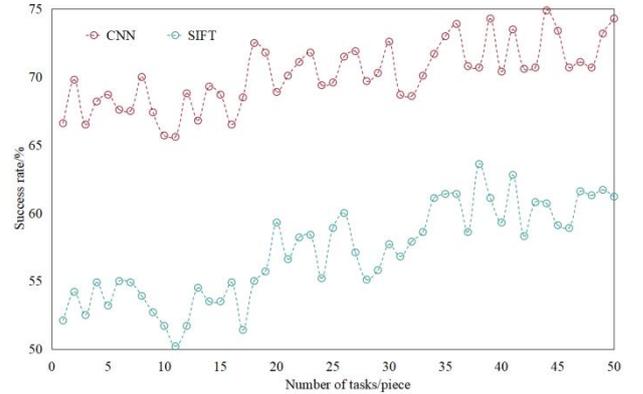

Fig. 2. Comparison of tracking success rates

Figure 3 provides a detailed comparison of the performance of our algorithm with traditional SIFT based object tracking algorithms in terms of recall rate, a key indicator. From the figure, it can be clearly seen that the algorithm proposed in this paper exhibits significant advantages in recall rate, mainly due to its unique algorithm design and implementation [26-28]. The key to achieve higher recall rate in this algorithm lies in its clever integration of correlated features and the advantages of CNN algorithm[29]. The powerful feature learning ability of CNN enables algorithms to automatically extract rich and discriminative feature representations from raw video frames, which are crucial for distinguishing targets from backgrounds and maintaining the continuity of targets between different frames [30-31]. At the same time, the introduction of correlated features further enhances the adaptability of the algorithm in complex tracking scenarios, enabling the algorithm to more accurately identify and recall tracked targets, and maintain stable tracking performance even in challenging situations such as occlusion, deformation, or rapid movement of targets[32].

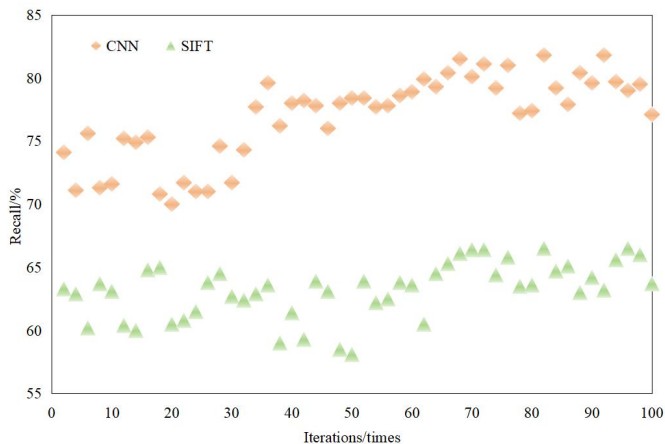

Fig. 3. Comparison of recall rates

Figure 4's detailed comparison highlights YOLOv3's significant advantage in time efficiency for object tracking over traditional SIFT-based methods and the emerging Transformer model. Traditional SIFT methods, while strong in feature description, suffer from slow processing due to complex extraction and matching[33]. The Transformer, despite its successes in NLP, struggles with computational efficiency and memory usage in object tracking[34]. Conversely, YOLOv3 integrates detection and recognition in a single-stage framework, leveraging deep CNN features, anchor boxes, and bounding box regression for efficient, accurate tracking[35]. By optimizing network structure and parameters, YOLOv3 markedly reduces computational complexity and time loss, while retaining high tracking accuracy[36].

To assess the real-time performance of the proposed algorithm, we compared its execution speed on GPU and CPU platforms. On GPU, the algorithm achieves 30 FPS, enabling smooth real-time tracking for video applications. However, on CPU, due to the increased computational load, the performance drops to 15 FPS. This difference highlights the significant speed-up that can be achieved by leveraging GPU acceleration, especially when handling computationally intensive tasks such as CNN-based video object tracking.

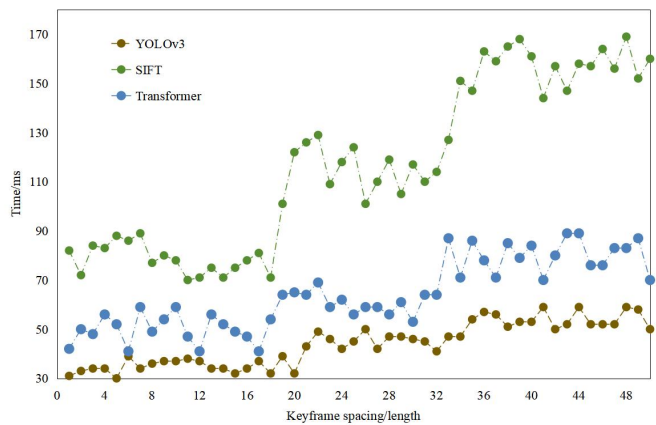

Fig. 4. Time loss comparison

Figure 5's in-depth comparison shows YOLOv3's notable advantages in loss function convergence speed for object tracking, outperforming traditional SIFT and Transformer methods. The rapid decline and stabilization of YOLOv3's loss during training indicate faster convergence. This stems from its multi-scale prediction, optimized network, and efficient loss function. Multi-scale prediction enhances detection robustness and accuracy; the optimized network cuts down unnecessary calculations and speeds up information processing. YOLOv3's loss function, balancing positioning, confidence, and classification errors, enables the model to quickly focus on key information, facilitating faster convergence[37-38].

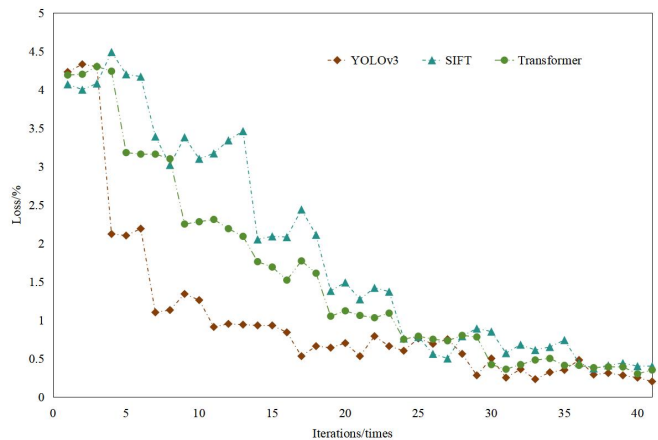

Fig. 5. Comparison of loss functions

IV. CONCLUSION

This paper designs and implements an efficient object tracking algorithm using CNN for complex scenarios. Experimental data shows its excellent performance in handling fast motion, partial occlusion, and complex backgrounds, significantly improving the tracking success rate. By integrating DL technology, it enhances real-time video processing with higher accuracy and stability, benefiting applications like intelligent monitoring and autonomous driving. The primary limitation of the proposed algorithm lies in its computational complexity, particularly when handling large-scale video streams in real-time applications. Future research will focus on reducing the computational complexity by integrating lightweight neural networks and exploring hardware acceleration techniques to enhance real-time performance in challenging environments.